\documentclass{article}
\usepackage{tabularx}
\usepackage{array}
\usepackage{setspace}
\usepackage{hyperref}
\usepackage{amsmath}
\usepackage{amsfonts}
\usepackage{graphicx}
\usepackage{placeins}
\usepackage{url}
\usepackage{float}
\usepackage{microtype}
\usepackage{graphicx}
\usepackage{subfigure}
\usepackage{booktabs} % for professional tables
\usepackage{tabularx}
\usepackage{amsmath}
\usepackage{amsfonts}
\usepackage{hyperref}

\usepackage[accepted]{arxiv}
\usepackage{amsmath}
\usepackage{amssymb}
\usepackage{mathtools}
\usepackage{amsthm}

\usepackage[capitalize,noabbrev]{cleveref}

\newcommand{\mcD}{\mathcal{D}}

\def \xb{\mathbf{x}}

\usepackage{tabularx}  % in your preamble

\theoremstyle{plain}

\theoremstyle{definition}

\theoremstyle{remark}

\usepackage[textsize=tiny]{todonotes}

\icmltitlerunning{An Attention-based Spatio-Temporal Neural Operator for Evolving Physics}

\begin{document}

\onecolumn
\icmltitle{An Attention-based Spatio-Temporal Neural Operator for Evolving Physics}




\begin{icmlauthorlist}
\icmlauthor{Vispi Karkaria}{nu}
\icmlauthor{Doksoo Lee}{nu}
\icmlauthor{Yi-Ping Chen}{nu}
\icmlauthor{Yue Yu}{lehigh}
\icmlauthor{Wei Chen}{nu}
\end{icmlauthorlist}

\icmlaffiliation{nu}{Northwestern University}
\icmlaffiliation{lehigh}{Lehigh University}

\icmlcorrespondingauthor{Wei Chen}{weichen@northwestern.edu}
\icmlcorrespondingauthor{Yue Yu}{yuy214@lehigh.edu}
%\icmlcorrespondingauthor{Anonymous Author}{Wei Chen (weichen@northwestern.edu)}
%\icmlcorrespondingauthor{Vispi Nevile Karkaria}{vispikarkaria2026@u.northwestern.edu}

\icmlkeywords{Machine Learning, ICML}

\vskip 0.3in



\printAffiliationsAndNotice{} %
%otherwise use the standard text.

\begin{abstract}
In scientific machine learning (SciML), a key challenge is learning unknown, evolving physical processes and making predictions across spatio-temporal scales. For example, in real-world manufacturing problems like additive manufacturing, users adjust known machine settings while unknown environmental parameters simultaneously fluctuate. To make reliable predictions, it is desired for a model to not only capture long-range spatio-temporal interactions from data but also adapt to new and unknown environments; traditional machine learning models excel at the first task but often lack physical interpretability and struggle to generalize under varying environmental conditions. To tackle these challenges, we propose the Attention-based Spatio-Temporal Neural Operator (ASNO), a novel architecture that combines separable attention mechanisms for spatial and temporal interactions and adapts to unseen physical parameters. Inspired by the backward differentiation formula (BDF), ASNO learns a transformer for temporal prediction and extrapolation and an attention-based neural operator for handling varying external loads, enhancing interpretability by isolating historical state contributions and external forces, enabling the discovery of underlying physical laws and generalizability to unseen physical environments. Empirical results on SciML benchmarks demonstrate that ASNO outperforms over existing models, establishing its potential for engineering applications, physics discovery, and interpretable machine learning.
\end{abstract}

\section{Introduction}

The increasing complexity of spatio-temporal data in scientific fields such as fluid dynamics, material science, and manufacturing, particularly in the context of Digital Twin technologies, has made the interpretability and accuracy of machine learning models critically important. Digital Twins—virtual replicas of physical systems—are becoming integral to modern manufacturing processes, enabling real-time monitoring, simulation, and optimization of physical assets \cite{kapteyn2021probabilistic, van2023digital, karkaria2024towards}. In such settings, the critical challenge lies in creating models that generalize to new physical parameters (e.g., PDE coefficients) while maintaining alignment with underlying physical principles to enable robust predictions and insights into system behavior. This dual capability enhances decision-making and facilitates system optimization by aligning predictions with real-world physics\cite{thelen2022comprehensive, karkaria2024optimization,chen2025real}. 

Existing methods often struggle to generalize across unseen physical environments, particularly when tasked with interpreting spatial and temporal dependencies in a separable yet synergistic manner. Neglecting this separability can obscure the physical mechanisms driving system behavior, resulting in less interpretable and reliable predictions \cite{rudin2022interpretable, molnar2020interpretable}. For example, in material modeling, observable phenomena such as deformation fields may provide limited information, and inferring hidden parameters such as material properties remains an ill-posed problem. Similarly, in physics-based applications like manufacturing, accurately capturing the interplay between spatial interactions and evolving temporal dynamics is essential for refining product quality and operational efficiency \cite{ko2022spatial}.

While existing data-driven surrogate architectures—such as transformer encoders, U-Net, DeepONet, and newer operator-based frameworks like the General Neural Operator Transformer (GNOT), Transolver, and Fourier Neural Operator (FNO) have been developed to capture spatio-temporal dependencies in physical systems \cite{lu2021learning, zhang2018human, hao2023gnot, wu2024transolver, wang2024p}, they often lack inherent mechanisms for reliable generalization to unseen conditions or for disentangling spatial and temporal influences. Transformers, despite their strength in modeling long-range temporal correlations via self-attention, frequently underrepresent the spatial interactions essential for nonlinear phenomena such as turbulent flows, reaction–diffusion processes, and phase transitions. Moreover, many of these approaches struggle to separate the roles of temporal evolution and spatial coupling, which limits their applicability in PDE-governed settings. Finally, their adaptability to systems with varying PDE coefficients remains limited, reducing their effectiveness across diverse physical scenarios. While the specific implementations vary—from sequence-based encoders to spectral and attention-driven operators—all of these methods share the common challenge of balancing expressivity with robust, physics-informed generalization.

To address these limitations, we propose the Attention-based Spatio-Temporal Neural Operator (ASNO). In ASNO, a transformer encoder captures temporal dynamics in the solution field, while an attention-based nonlocal operator handles the long-range spatial dependencies and the interplay between solution and loading/environment fields. Inspired by the implicit-explicit (IMEX) scheme, ASNO leverages the backward differentiation formula (BDF) to separate temporal effects from spatial loading/solution interactions. This separable architecture enhances generalizability to unseen physical parameters (e.g., initial conditions, loadings, environments) while improving interpretability.

The key contributions of this paper are as follows:
\begin{itemize}
    \item We introduce ASNO, a novel architecture inspired by the IMEX scheme, which discovers separate mechanisms for capturing temporal effects and spatial loading/solution interactions. 
    \item ASNO discovers a spatio-temporal relationship from data, enabling zero-shot generalizability to PDEs with unseen initial conditions, loadings, and environments.
    \item The attention mechanism provides insights into the separable contributions of temporal and spatial dynamics, overcoming the limitations of models focused on only one of the two.
    \item We conduct experiments on examples ranging from chaotic systems and PDE solving problems to real-world applications. Results demonstrate the advantages of ASNO over baseline transformer models and neural operator models in terms of generalizability, long-term stability, and interpretability.
\end{itemize}

%The rest of this paper is structured as follows: Section 2 reviews background and related work on transformer-based models and neural operators. Section 3 introduces the ASNO architecture, detailing the Transformer Encoder and Nonlocal Attention Operator components, along with a theoretical convergence analysis of ASNO’s attention mechanisms. Section 4 presents empirical results on benchmarks, including the Lorenz system, PDE-based simulations, and a case study in additive manufacturing. Finally, Section 5 concludes with a summary, implications, and future research directions.

% The key contributions of this paper are as follows:
% \begin{itemize}
%     \item We propose a novel architecture, ASNO, that effectively integrates transformer-based temporal modeling with NAO, significantly improving the capture of long-range spatial and temporal dependencies.
%     \item Our model facilitates detailed spatial and temporal analysis, overcoming the limitations of methods that focus solely on one of these aspects, and enabling accurate modeling of intricate spatio-temporal systems.
%     \item ASNO solves both forward (physics prediction) and inverse (physics discovery) problems in complex spatio-temporal systems, such as PDE-based simulations and chaotic systems.
%     \item Our model demonstrates enhanced predictive accuracy, computational efficiency, and scalability, making it suitable for real-time applications like Digital Twins and advanced manufacturing systems.
% \end{itemize}

\section{Background and Related Work}

\subsection{Transformers for Time Series Data}

Traditional time series models such as AutoRegressive Integrated Moving Average (ARIMA) and Long Short-Term Memory (LSTM) have effectively captured short-term dependencies but face limitations when dealing with long-range dependencies and highly non-linear, multivariate data \cite{nelson1998time, graves2012long}. Transformers, initially developed for natural language processing (NLP), revolutionized time series modeling by introducing self-attention mechanisms that capture dependencies across long sequences \cite{vaswani2017attention, tang2021probabilistic, zerveas2021transformer, zhou2021informer}. This attention mechanism allows transformers to weigh different parts of the input sequence based on their relevance, making them highly suitable for long-sequence tasks \cite{zhao2024tfformer}.

Despite their strengths, transformers can be computationally expensive. Their complexity grows quadratically with sequence length, limiting scalability, especially when applied to large datasets \cite{fournier2023practical}. This has driven the development of models like Informer \cite{zhou2021informer} and Reformer \cite{kitaev2020reformer}, which employ sparse attention mechanisms to reduce complexity. Models such as the Temporal Fusion Transformer (TFT) have improved upon traditional transformers by using attention scores to enhance interpretability and effectively address practical challenges like variable input lengths and missing data \cite{lim2021temporal}; these features are byproducts of TFT’s core innovations in handling complex temporal dependencies.

These developments underscore the adaptability of transformer architectures in handling diverse temporal tasks, though they remain limited in their ability to generalize to new physical parameters (e.g., PDE coefficients) and lack mechanisms that explicitly separate spatial and temporal interactions for interpretability. \cite{cheng2024mamba}. 

\subsection{Neural Operators in Scientific Machine Learning}

Neural operators have become essential in scientific machine learning, particularly for modeling mappings between function spaces \cite{li2020fourier}. These models are effective at solving forward problems in physics-based systems, such as those governed by PDEs, by serving as efficient surrogates for physical systems and offering black-box approximations without explicitly interpreting the underlying physical laws \cite{kovachki2023neural, azizzadenesheli2024neural, cao2024laplace, wen2022u}. For instance, DeepONet utilizes separate networks to encode input functions and evaluation points, thereby learning the nonlinear operator that maps function inputs to output values at arbitrary spatial locations. \cite{lu2021learning, lu2019deeponet}. However, as the problem's dimensionality increases, DeepONet faces scalability issues \cite{mandl2024separable}. Similarly, Fourier Neural Operator (FNO), which operates in the frequency domain to better capture global spatial dependencies, becomes computationally expensive either at higher resolutions or when handling non-smooth data \cite{li2020fourier}. Recent advancements, such as the GNOT and Transolver, have improved the flexibility of neural operators for handling complex, spatially distributed data. GNOT incorporates heterogeneous normalized attention layers and geometric gating mechanisms, enabling it to address multi-scale problems and integrate diverse data sources effectively \cite{hao2023gnot}. Similarly, Transolver leverages transformer-based architectures to solve PDEs on irregular geometries, enhancing its adaptability to various applications. However, these models often struggle to generalize to unseen PDE parameters (e.g., new initial conditions or environments) and lack interpretability in disentangling spatial and temporal interactions \cite{wang2024p, wu2024transolver}.

Most neural operators primarily address forward PDE problems—predicting future states from current conditions by capturing spatio-temporal dependencies—but struggle with inverse problems, like reconstructing unknown PDE parameters, which often require prior knowledge of governing equations and are generally ill-posed \cite{molinaro2023neural}. Integrating forward and inverse problem-solving capabilities into a single spatio-temporal framework enables the development of more versatile models that can not only predict system behaviors but also infer hidden states and causal mechanisms. This dual functionality enhances model robustness and adaptability—qualities vital for scientific applications where elucidating system dynamics, which may inherently encode underlying causes, contributes to more informed simulation, and potentially supports control and optimization efforts, even if these are ultimately executed through computational rather than interpretative means \cite{kim2024review}. 

\section{Model Architecture}

This section presents ASNO, a spatio-temporal neural operator integrating the Transformer Encoder for temporal dependencies and the nonlocal attention mechanism for spatial interactions. We outline the Transformer Encoder, followed by the nonlocal attention operator model, and explain their combined role in the form of a backward differentiation formula (BDF) for complex spatio-temporal prediction. 
%In empirical experiments of Section \ref{sec:exp}, we investigate the convergence properties of ASNO’s attention mechanisms, where the integration of the NAO and Transformer Encoder ensures stable, accurate long-term predictions crucial for scientific applications.

\subsection{Backward Differentiation Formula (BDF)}

Backward differentiation formula (BDF) is a family of popular numerical schemes for stiff differential equations, thanks to its high-order accuracy and large region of absolute stability \cite{fredebeul1998bdf}. For a initial-value differential equation:
$$\dot{X}=F(t,X),\; X(t_0)=X_0,$$
the general formula for a BDF can be written as
\begin{equation}\label{eq:BDF}
\sum_{k=1}^n \alpha_k X_{m-k+1}+X_{m+1} = \Delta t \beta F((m+1)\Delta t,X_{m+1}).
\end{equation}
Here, $\Delta t$ denotes the time step size, and $X_{m}$ is the approximated solution at time instance $t_0+m\Delta t$. Notice that \eqref{eq:BDF} can be naturally decomposed as the two-phase implicit-explicit (IMEX) scheme \cite{ascher1995implicit}:
\begin{align}
&-\sum_{k=1}^n \alpha_k X_{m-k+1}=\tilde{X}_{m+1},\label{eq:bdf_ex}\\
&\tilde{X}_{m+1}-{X}_{m+1}+\Delta t \beta F((m+1)\Delta t,X_{m+1})=0.\label{eq:bdf_im}
\end{align}

In the explicit step \eqref{eq:bdf_ex}, $n$ historical steps are accumulated, hence it characterizes the temporal interactions. Notice that when the system is homogeneous, i.e., $F\equiv 0$, we have $\tilde{X}_{m+1}={X}_{m+1}$. That means, the explicit steps generate an estimated temporal extrapolated solution for a system without external loading $F$. On the other hand, the implicit step \eqref{eq:bdf_im} takes $\tilde{X}_{m+1}$ as the input and tries to solve a (possibly nonlinear) static equation of $X_{m+1}$. Hence, this step is characterized by capturing the spatial interaction between components of $X$, as well as their interplay with the external loading $F$.

Based on this idea, our key innovation in ASNO is to design a separable architecture that captures the temporal and spatial effects respectively. In particular, our architecture resembles the implicit-explicit decomposition of BDF. We notice that the external loading $F$ is non-autonomous, meaning that it depends not only on $X$ but also on another (hidden) time variant quantity, which can be seen as the hidden system state, $S_{m+1}$, to be discovered through learning.

We now formally state the learning settings in ASNO: consider multiple physical systems which can be described by a series of initial-valued problems:
\begin{equation}
\dot{X}^{(\eta)}=F(S^{(\eta)},X^{(\eta)}),\; X^{(\eta)}(t_0)=X^{(\eta)}_0,
\end{equation}
where $\eta=1,\cdots,S$ is the sample/system index. Then, for each system, we assume a sequence of observations:
\begin{equation}
\mcD^{(\eta)}=\{X^{(\eta)}(m\Delta t),F^{(\eta)}(m\Delta t)\}_{m=0}^{T/\Delta t}
\end{equation}
are provided. The goal of ASNO are three folds:
\begin{enumerate}
\item Identify a temporal extrapolation rule \eqref{eq:bdf_ex} from data, for a stable and accurate long-term prediction of $X$.
\item For each system with hidden state $S^{(\eta)}$, infer the underlying context or structure of $S^{(\eta)}$ directly from observational data, without relying on labeled supervision or prior knowledge of the state.
\item For each new and unseen system with the first $n$ steps of $(X,F)$ given, provide a zero-shot prediction rule without training.
\end{enumerate}

\subsection{Explicit Step: Temporal Extrapolation}

\begin{figure*}[h!]
    \centering
    \includegraphics[width=0.9\linewidth]{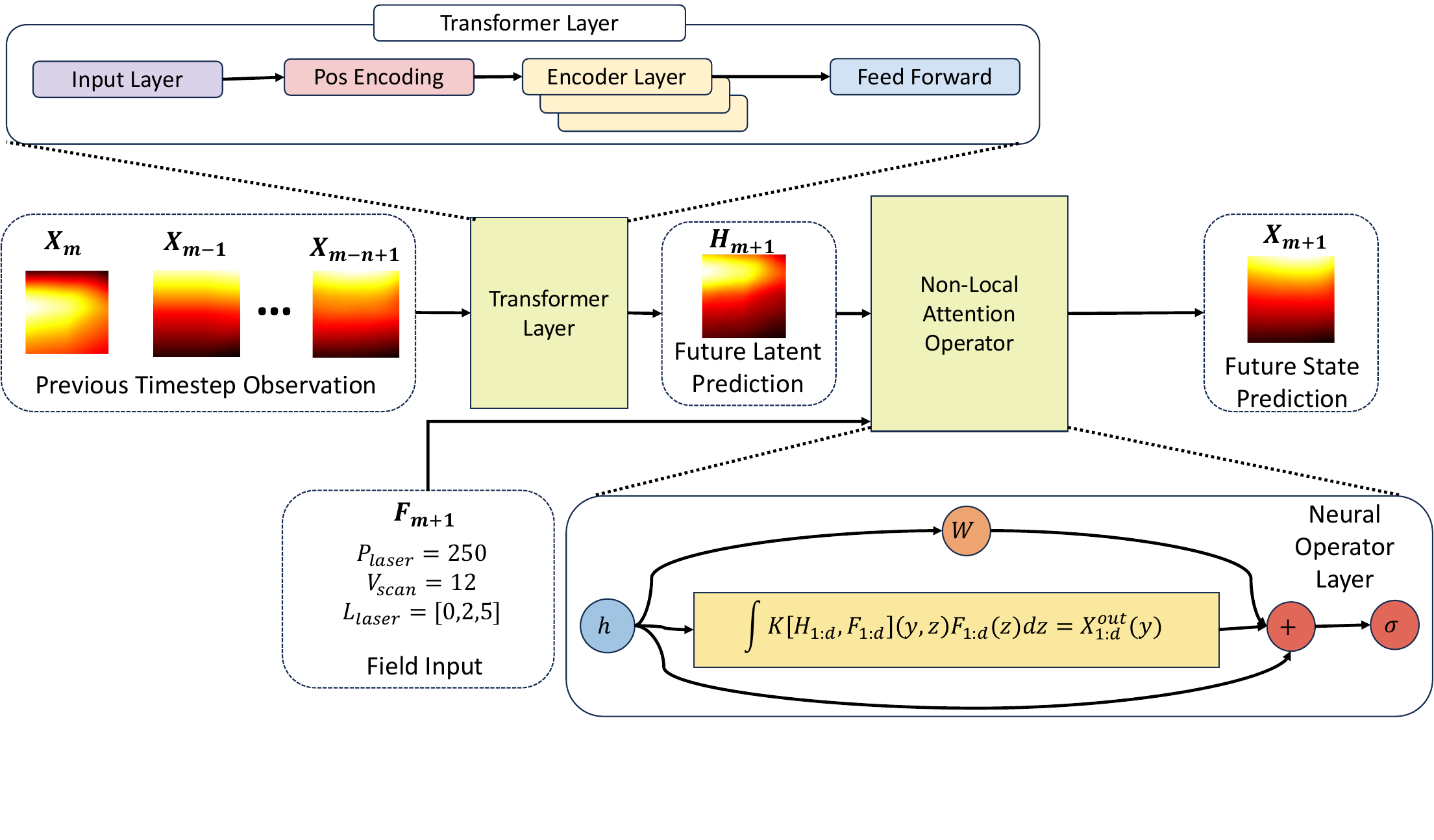} % Reduced width slightly
    \vspace{-1.5cm}  % Reduce the vertical spacing below the figure for a tighter fit
    \caption{The ASNO framework with the Transformer Encoder for capturing temporal dependencies and the Non-Local Attention Operator (NAO) for modeling spatial interactions, following a Backward Differentiation Formula (BDF) approach.}
    \label{fig:ASNO_framework}
\end{figure*}

As the first component of ASNO, we employ a Transformer Encoder to resemble the explicit step \eqref{eq:bdf_ex}. In particular, it processes time series data by capturing long-range temporal dependencies, and predicts future states of systems governed by high-dimensional, nonlinear dynamical equations.

In Transformer Encoder, the input to the encoder is a sequence of observations from previous time steps, denoted as \(\{X_m, X_{m-1}, X_{m-2}, \dots, X_{m-n+1} \} \), where \( n \) represents the number of past time steps considered. Each observation \( X_{t} \in \mathbb{R}^{d} \) is projected into an embedding space of dimension \( d_{\text{embed}} \), with positional encodings added to preserve temporal order:
{\small
\begin{equation}
\begin{aligned}
    E_{t} = X_{t} W_E + P_{t},
\end{aligned}
\end{equation}
}

where \( W_E \in \mathbb{R}^{d \times d_{\text{embed}}} \) is the embedding matrix, and \( P_{t} \in \mathbb{R}^{d_{\text{embed}}} \) is the positional encoding vector. The encoded sequence \( \{E_{m}, E_{m-1}, \dots, E_{m-n+1}\} \) is then passed through multiple layers of the transformer encoder, each consisting of a multi-head self-attention mechanism and a feed-forward network, allowing the model to learn complex dependencies across time steps.

In the attention mechanism \cite{niu2021review}, the query, key, and value matrices are derived from the encoded input sequence using weights \(W_{Tq}, W_{Tk}, W_{Tv} \in \mathbb{R}^{d_{\text{embed}} \times d_t}\), which are learnable parameters for the query, key, and value transformations, respectively.
{\small
\begin{equation}
\begin{aligned}
    Q = E_t\,W_{Tq}, \quad K = E_t\,W_{Tk}, \quad V = E_t\,W_{Tv},
\end{aligned}
\end{equation}
}
The attention score between the query at the \(t\)-th time step, \( Q_t \), and the keys \( K \) from previous steps is calculated as a weighted sum, capturing dependencies between the current time step and previous steps. Formally, for the \(m\)-th time step, the future latent space \( H_{m+1} \) is predicted using the attention-weighted sum of past values:
{\small
\begin{equation}
\begin{aligned}
    H_{m+1} = \text{TE}(X_{m}, X_{m-1}, \dots, X_{m-n+1}) = \sum_{i=1}^{n} \alpha_{m,i} \cdot V_{m},
\end{aligned}
\label{eq:H}
\end{equation}
}
where \(\alpha_{m,i}\) represents the attention weights between \(Q_m\) and \(K_{m-i+1}\), computed as
{\small
\begin{equation}
\begin{aligned}
    \alpha_{m,i} = \frac{\exp\!\bigl(\tfrac{Q_m \cdot K_{m-i+1}^\top}{\sqrt{d_t}}\bigr)}%
                           {\sum_{j=1}^{n}\exp\!\bigl(\tfrac{Q_m \cdot K_{m-j+1}^\top}{\sqrt{d_t}}\bigr)},
\end{aligned}
\end{equation}
}
where \(d_t\) is the dimensionality of the key vectors. This enables the transformer to focus on the most relevant past steps when extrapolating the latent state.

The predicted latent variable \(H_{m+1}\in\mathbb{R}^{d_{\text{embed}}}\) corresponds to the homogeneous BDF extrapolation \(\tilde{X}_{m+1}\) and is passed into the NAO for the implicit spatial correction step \eqref{eq:bdf_im}.

\subsection{Implicit Step: Static PDE Solver}

For the implicit step \eqref{eq:bdf_im}, we take $\tilde{H}_{m+1}$ and $F_{m+1}$ as inputs, and aim to approximate the solution $X_{m+1}$. Since \eqref{eq:bdf_im} can be seen as a nonlinear static PDE solver, we propose to learn a neural operator as the surrogate solution operator. However, due to the possible change of underlying system state, $S$, generic neural operators \cite{lu2019deeponet,li2020fourier,you2022learning} would fail this task because they focus on the solution of one single PDE. Herein, we propose to employ the attention mechanism, in the form of the Nonlocal Attention Operator (NAO) \cite{yu2024nonlocal}, which combines attention and neural operator learning, offering a generalizable solution operator across different PDEs. In particular, NAO expands on traditional attention by effectively modeling long-range spatial dependencies crucial for capturing continuous, nonlocal interactions within physical systems \cite{yu2024nonlocal}. This approach allows NAO to aggregate global context, addressing limitations of standard attention in handling nonlocal, spatially distributed information, especially beneficial for high-dimensional data processing in fluid dynamics and thermodynamics.

Formally, NAO is designed to solve both forward and inverse problems, with the underlying system dynamics encapsulated by operators that map input functions $h \in \mathcal{H}$ to output functions $g \in \mathcal{F}$. This operator is defined as:
{\small
\begin{equation}
L_K[h] + \epsilon = g,
\end{equation}
}
where $\mathcal{H}$ and $\mathcal{F}$ are Banach spaces, $K$ is the nonlocal interaction kernel, and $\epsilon$ denotes additive noise accounting for model–system discrepancies. NAO introduces a critical component: a kernel map constructed via attention. This kernel map estimates the nonlocal kernel based on both input \(h\) and output \(g\), enabling the model to capture global context across physical states. 

Herein, we take $h$ as the latent variable field together with the loading fields, and $g$ as the prediction of the next time step. The latent variable and loading fields are discretized over the spatial mesh \(\{y_k\}_{k=1}^N\) as:
{\small
\begin{equation}
\begin{aligned}
{H}_{1:d} &= \bigl({H}_j(y_k)\bigr)_{1 \leq j \leq d,\; 1 \leq k \leq N}, \\
F_{1:d} &= \bigl(F_j(y_k)\bigr)_{1 \leq j \leq d,\; 1 \leq k \leq N},
\end{aligned}
\end{equation}
}
with $d$ being the feature size, which is usually as $d$ steps from the same system.

The overall process in NAO iteratively transforms an initial state 
\[
J_0 = \bigl({H}_{1:d},\,F_{1:d}\bigr)
\]
through \(T\) steps, each applying attention with residual connections:
{\small
\begin{equation}
J_t 
= \mathrm{Attn}\bigl(J_{t-1};\theta_t\bigr)\,J_{t-1} \;+\; J_{t-1}:=(J_t,F_t),\quad 1\le t \le T,
\end{equation}
}
where
{\small
\begin{equation}
\mathrm{Attn}[J;\theta_t]
=\sigma\!\Bigl(\tfrac{1}{\sqrt{d_k}}\,J\,W_{Q_t}\,(W_{K_t})^\top\,J^\top\Bigr).
\end{equation}
}
After \(T\) steps, we form the learned kernel via a kernel map:
\small{
\begin{equation}
\begin{aligned}
K\bigl[H_{1:d},\,F_{1:d};\,\theta\bigr]
=&\;W_{P,h}\,\sigma\!\Bigl(\tfrac{1}{\sqrt{d_k}}\,(J_{T})^\top\,W_{Q_{T+1}}\,(W_{K_{T+1}})^\top\,J_{T}\Bigr)\\
&+W_{P,f}\,\sigma\!\Bigl(\tfrac{1}{\sqrt{d_k}}\,(F_{T})^\top\,W_{Q_{T+1}}\,(W_{K_{T+1}})^\top\,J_{T}\Bigr),
\end{aligned}
\label{eqn:K}
\end{equation}
}
and compute the output
{\small
\begin{equation}
X^{\mathrm{out}}_{1:d}(y)
=\int K\bigl[{H}_{1:d},\,F_{1:d}\bigr](y,z)\,F_{1:d}(z)\,\mathrm{d}z.
\end{equation}
}
In this architecture, \(\theta=\{W_{P,h},W_{P,f},W_{Q_t},W_{K_t}\}\) are trainable, \(W_{Q_t},W_{K_t}\) are query/key matrices, \(\sigma\) is a linear activation function following the suggestions in \cite{yu2024nonlocal,cao2021choose,lu2025transformer}. To optimize $\theta$, we minimize the training loss across all \(S\) systems:
{\small
\begin{equation}
\mathcal{L}
=\sum_{\eta=1}^S
\Bigl\|\int K[{H}_{1:d},\,F_{1:d}](y,z)\,F_{1:d}(z)\,\mathrm{d}z
-X^{out,true}_{1:d}(y)\Bigr\|_{L^2}^2.
\end{equation}
}
The NAO thus models nonlocal spatial interactions via a data‐dependent kernel on the latent space \(H\), capturing complex system states \(S\) across diverse PDEs and enhancing generalizability.

\subsection{ASNO: Summary and Algorithm}

The ASNO architecture combines the Transformer Encoder and the NAO to effectively capture both temporal and spatial dependencies in complex spatio-temporal data, as shown in Figure \ref{fig:ASNO_framework}. 

\begin{figure*}[h!]
    \centering
    \includegraphics[width=0.9\linewidth]{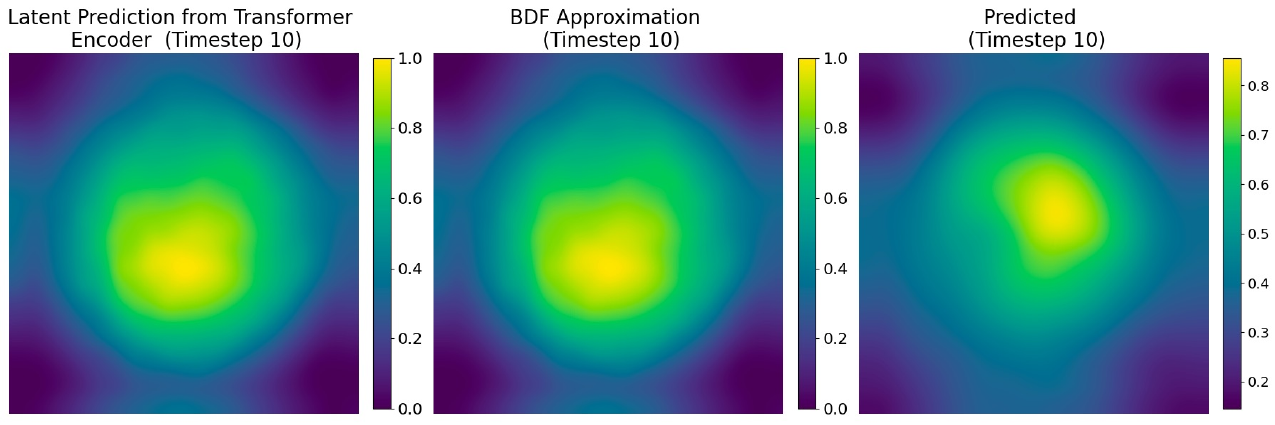}
    \vspace{-15pt}
    \caption{Comparison of the latent prediction from the Transformer Encoder (\(H_{m+1}\)), the BDF approximation (\(\tilde{X}_{m+1}\)), and the ASNO-predicted output at Timestep 10. The latent prediction and BDF approximation closely align, and the final ASNO output differs as the NAO adjusts the prediction using field-based corrections.}
    \vspace{-10pt}
    \label{fig:BDF_vs_Latent}
\end{figure*}

As the Transformer Encoder processes the past observations and outputs the latent variable \(H_{m+1}\) using Equation \ref{eq:H}, \(H_{m+1}\) is fed into the NAO to provide the temporal information. This process can be represented as:
\begin{equation}
\begin{aligned}
    X^{out}_{m+1} &= \mathrm{NAO}\bigl(H_{m+1},\,F_{m+1}\bigr) \\
    &= \mathrm{NAO}\bigl(\mathrm{TE}(X_{m}, X_{m-1}, \dots, X_{m-n+1}),\,F_{m+1}\bigr) \\
    &= \mathrm{ASNO}(X_{m}, X_{m-1}, \dots, X_{m-n+1},\,F_{m+1}).
\end{aligned}
\end{equation}

Since the temporal information is compressed into a latent representation by the Transformer Encoder, and the NAO leverages it to learn spatial relationships, the combination of the two modules allows ASNO to learn spatio-temporal dependencies. The Transformer Encoder and NAO are jointly trained with a loss function that optimizes the model's ability to capture complex dependencies efficiently, making it adaptable to a wide range of scientific machine-learning tasks.

\section{Experiments}

In this section, we evaluate ASNO's performance on case studies involving chaotic systems and spatio-temporal PDEs, each chosen to illustrate its strengths in capturing complex dynamical behaviors crucial to scientific machine learning.

\subsection{Learning a PDE Solution Operator for Darcy Flow}

We evaluate ASNO on a dynamic Darcy flow equation, which models fluid flow through porous media and exhibits complex spatio‐temporal dynamics. On a square domain $\Omega:=[0,1]^2$, The governing equation writes:
\begin{align}
\label{eqn:darcy}
\dfrac{\partial p(t,x)}{\partial t}-\nabla\cdot(b(x)\nabla p(t,x))=g(t,x),\;\quad x\in \Omega,\qquad p(x)=0,\quad\xb\in\partial \Omega.
\end{align}
In this context, we aim to learn the solution operator of Darcy's equation and compute the pressure field $p(t,x)$, given the source field $g(t,x)$. The training dataset consists of 100 time‐series profiles on a \(21\times21\) grid, each with 100 timesteps. Here, each profile is determined by a different hidden microstructure field $b(x)$. Input–output pairs are generated via a sliding window of five consecutive timesteps (i.e., stride 1), yielding 96 samples per profile. We apply 20 random permutations per trajectory, resulting in 153{,}600 training samples and 38{,}400 test samples (80:20 split). All baseline models (FNO, U-Net, Transolver, GNOT, DeepONet, Transformer Encoder, Transformer Encoder + NAO) use the same inputs: the past five states \(X_{m-4},\dots,X_m\) and the current forcing field \(F_m\), where the \emph{state} \(X_m\) represents the pressure field at time step \(m\) and the \emph{forcing field} \(F_m\) encodes the source or sink terms influencing fluid flow at each spatial location.

Table~\ref{tab:darcy_performance_fullparams} compares the number of trainable parameters, GPU memory usage, test loss on the standard test set, and out-of-distribution (OOD) test losses for two variants of OOD data. Here, OOD refers to inputs whose underlying distributions differ from those seen during training. Assessing OOD performance is important because it measures a model’s generalizability under distributional shifts. In our case, the OOD-f set uses time-varying source fields drawn from a different Gaussian random field (GRF) distribution than the training set. Specifically, while training samples use source functions \(g(t, x)\) drawn from a GRF with fixed hyperparameters \(\alpha = 2\) and \(\tau = 3\), the OOD-f samples are generated by modulating these GRFs with a sinusoidal time component and scaling their amplitudes by a factor of 200. In particular, the source term evolves over time as \(g(t, x) = \tilde{g}(x) \cdot \sin(t)\), where \(\tilde{g}(x)\) is the spatial GRF realization and \(t = q \Delta t\) is the temporal index with \(\Delta t = 5 \times 10^{-5}\). This construction introduces a dynamic temporal structure not present in the training distribution, thereby altering both the spatial and temporal characteristics of the source field. Meanwhile, the OOD-b set modifies the spatial distribution of the permeability field \(b(x)\), which is a crucial hidden input in Darcy flow. While the training set uses binary bi-phase microstructures generated from a GRF with parameters \(\alpha_{\chi} = 4\) and \(\tau_{\chi} = 5\), the OOD-b set uses a rougher, higher-frequency microstructure generated from a GRF with \(\alpha_{\chi} = 7\) and \(\tau_{\chi} = 6\). This change leads to sharper, more fragmented phase boundaries and higher heterogeneity in the underlying permeability, directly impacting the operator’s ability to generalize to previously unseen flow geometries. Generalizability is quantified by the extent to which the test loss increases on these OOD sets compared to the standard test set: smaller increases imply better robustness. ASNO achieves the lowest best and OOD test losses with a comparative number of trainable parameters among the neural-operator baselines, indicating the strongest OOD generalization with respect to both types of distributional shifts.

\begin{table}[h!]
\centering
\footnotesize
\renewcommand{\arraystretch}{1.2}
\setlength{\tabcolsep}{4pt}
\caption{Performance of different models on Darcy flow and Darcy OOD test sets.}
\label{tab:darcy_performance_fullparams}
\begin{tabularx}{\linewidth}{
  >{\centering\arraybackslash}X |
  >{\centering\arraybackslash}X |
  >{\centering\arraybackslash}X |
  >{\centering\arraybackslash}X |
  >{\centering\arraybackslash}X |
  >{\centering\arraybackslash}X
}
\hline\hline
\textbf{Model} 
  & \textbf{Trainable Params} 
  & \textbf{GPU (MB)} 
  & \textbf{Best Test Loss} 
  & \textbf{Best OOD-f} 
  & \textbf{Best OOD-b} \\
\hline
ASNO        & 760,234    & 181   & 0.0368 & \textbf{0.0673} & \textbf{0.0982} \\
FNO         & 900,224    & 214   & 0.0768 & 0.1129          & 0.1892          \\
U-Net       & 820,994    & \textbf{123} & 0.1150 & 0.1523          & 0.2224          \\
Transolver  & 810,573    & 422   & 0.0428 & 0.0721          & 0.1535          \\
GNOT        & 760,349    & 208   & 0.0516 & 0.0811          & 0.1729          \\
DeepONet    & 6,230,000 & 2146 & 0.0537 & 0.0826          & 0.1826          \\
Trans. Enc. & 1,620,394 & 173  & 0.0559 & 0.0927          & 0.1736          \\
Linear Enc. + NAO   & \textbf{720,398} & 165 & 0.05474 & 0.1245 & 0.1394 \\
\hline\hline
\end{tabularx}
\end{table}

ASNO not only provides a high‐performing predictor but also reveals underlying physical mechanisms through its learned representations. Its separable design helps to learn the implicit–explicit decomposition of backward differentiation formulas. Moreover, its modular structure enhances interpretability by allowing independent analysis of temporal and spatial contributions to the final prediction. To illustrate this, we compared the temporal predictions of the Transformer Encoder component with a classical BDF5 numerical approximation:

\begin{align}
\tilde{X}_m &= \frac{12}{137}\,X_{m-5}
             - \frac{75}{137}\,X_{m-4}
             + \frac{200}{137}\,X_{m-3}
             \nonumber\\
           &\quad - \frac{300}{137}\,X_{m-2}
             + \frac{300}{137}\,X_{m-1}\,,
\label{eq:bdf5}
\end{align}

showing that the learned latent representation \(H_m\) closely matches this high‐order temporal extrapolation (see Figure~\ref{fig:BDF_vs_Latent}). For the spatial interactions captured by the NAO, we compared the learned kernel with a theoretical kernel derived from the Darcy discretization. Starting from

\begin{align}
\tilde{X}_m &= X_m + \frac{60}{137}\,\Delta t\,\bigl(A\,X_m + F_m\bigr)\,,
\label{eq:spatial_expl}
\end{align}

one obtains

\begin{align}
X_m &= \Bigl(I + \tfrac{60}{137}\,\Delta t\,A\Bigr)^{-1}
      \Bigl(\tilde{X}_m - \tfrac{60}{137}\,\Delta t\,F_m\Bigr)\,,
\label{eq:spatial_impl}\\
K_{\mathrm{true}} &= -\frac{60}{137}\,\Delta t\;
    \Bigl(I + \tfrac{60}{137}\,\Delta t\,A\Bigr)^{-1}\,.
\label{eq:true_kernel}
\end{align}

\begin{figure*}[h!]
    \centering
    \includegraphics[width=\linewidth]{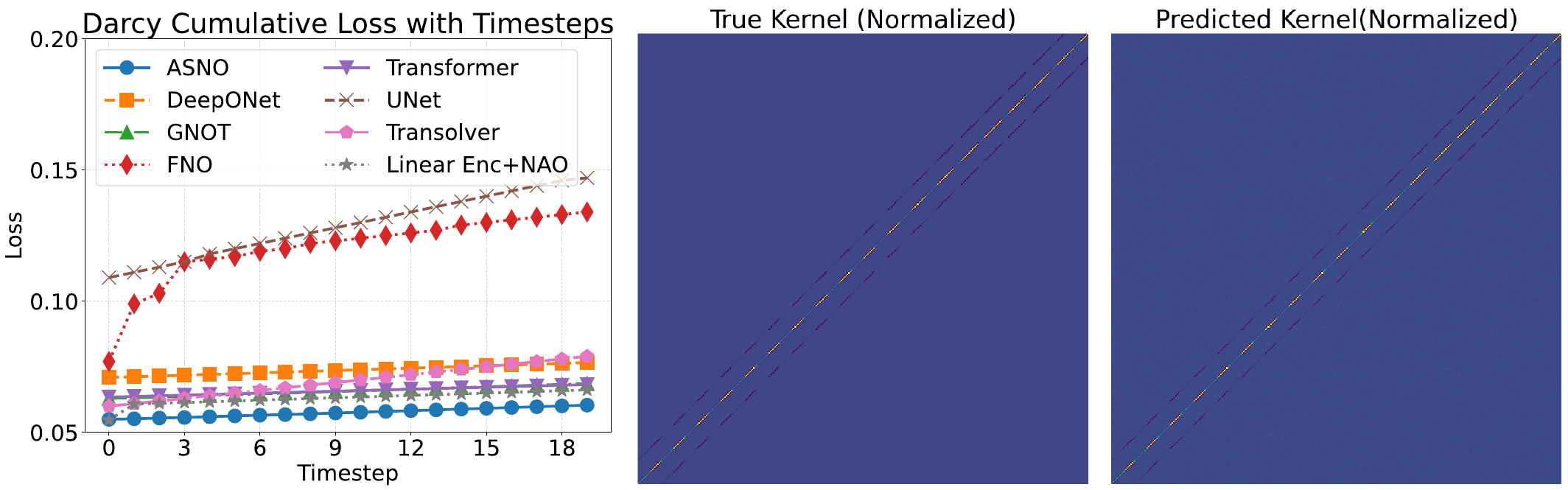}
    \vspace{-10pt}
    \caption{Left: Cumulative error over time for various models on the Darcy flow benchmark, illustrating ASNO’s long-term stability. Right: Comparison between the ground-truth Darcy kernel and the spatial kernel learned by the Nonlocal Attention Operator (NAO), demonstrating alignment with the underlying physics.}
    \label{fig:nao_kernel}
\end{figure*}

Figure \ref{fig:nao_kernel} compares the ground-truth kernel and the learned kernel from the NAO component, showing good agreement between the two. Additionally, the cumulative loss over time for various models illustrates that NAO outperforms existing methods in accuracy. When computing long-term time integration, each prediction \(X_t^{\mathrm{pred}}\) is fed back as an input for the next step, so errors accumulate over successive rollouts. The cumulative loss—also referred to as long-term time integration —is computed as

\begin{equation}
E_T \;=\; \sum_{t=1}^{T} \bigl\|X_t^{\mathrm{true}} - X_t^{\mathrm{pred}}\bigr\|_{L^2}\,,
\label{eq:cumulative_error}
\end{equation}

where \(X_t^{\mathrm{true}}\) and \(X_t^{\mathrm{pred}}\) denote the ground-truth and predicted states at timestep \(t\). This confirms that the spatial mechanism captured by NAO effectively reflects the physical structure of the underlying PDE and maintains strong predictive performance across timesteps.

These results show that ASNO balances predictive accuracy, interpretability, and computational cost, and maintains stable performance even under variations in forcing and permeability.

\subsection{Chaotic ODE: Lorenz System}

In this section, we evaluate the performance of ASNO on the Lorenz system, a well-known nonlinear dynamical model featuring high sensitivity to initial conditions:
\begin{align}
\frac{\partial x}{\partial t} &= \sigma (y(t) - x(t))+g_1(t), \label{eq:lorenz1} \\
\frac{\partial y}{\partial t} &= x(t)(\rho - z(t)) - y(t)+g_2(t), \label{eq:lorenz2} \\
\frac{\partial z}{\partial t} &= x(t)y(t) - \beta z(t)+g_3(t). \label{eq:lorenz3}
\end{align}
The dataset consists of 2000 time-series profiles with $100$ different coefficients combinations ($\sigma,\rho,\beta$), $20$ different loading functions $(g_1,g_2,g_3)$, and the same initial conditions $(x(0),y(0),z(0))=(0,1,0)$. Each profile contains 1000 timesteps. 80\% profiles are used for training and 20\%  are for testing. To construct input–output pairs, we apply a sliding window of length 5 that moves at every timestep, yielding 996 windows per profile (from \(t=1\) through \(t=996\)); we discard the final partial window to obtain 995 samples per profile. To improve model generalization and robustness, we augment the data by permuting the order of entire 5-step windows within each trajectory, rather than shuffling individual timesteps, which preserves local temporal structure while increasing data diversity, and by randomly permuting the profile indices to avoid overlap between train and test sets. This results in a total of \(80\times995\times20=1{,}592{,}000\) training samples and \(20\times995\times20=398{,}000\) testing samples without leakage. ASNO is compared against five baselines (Transolver, DeepONet, Transformer Encoder, GNOT, and Linear Model + NAO), all trained under identical input formats: each model receives the five previous states \(\{X_{m-4},\dots,X_m\}\) as well as the current forcing \(F_m\). We did not include architectures such as U-Net, FNO, and GNOT in our Lorenz comparison due to architectural mismatch: U-Net is designed for high-dimensional spatial grids and relies on convolutional down-/up-sampling; applying it to a low-dimensional ODE time series forces unnatural reshaping and often leads to over-smoothing of the rapid state changes characteristic of chaos. FNO similarly assumes a spatial Fourier basis—its global spectral filters excel for PDEs on regular meshes but provide little benefit (and much overhead) when modeling three-dimensional trajectories in \((x,y,z)\). GNOT builds in geometric gating layers that require constructing a graph or manifold structure even for these 3-variable ODEs, adding implementation complexity and extra hyperparameters with little payoff. The results in Table~\ref{tab:chaotic_system_results} show that ASNO achieves the lowest test loss (0.000794), demonstrating its predictive accuracy on this chaotic system.  

\begin{figure*}[h!]
    \centering
    \includegraphics[width=\linewidth]{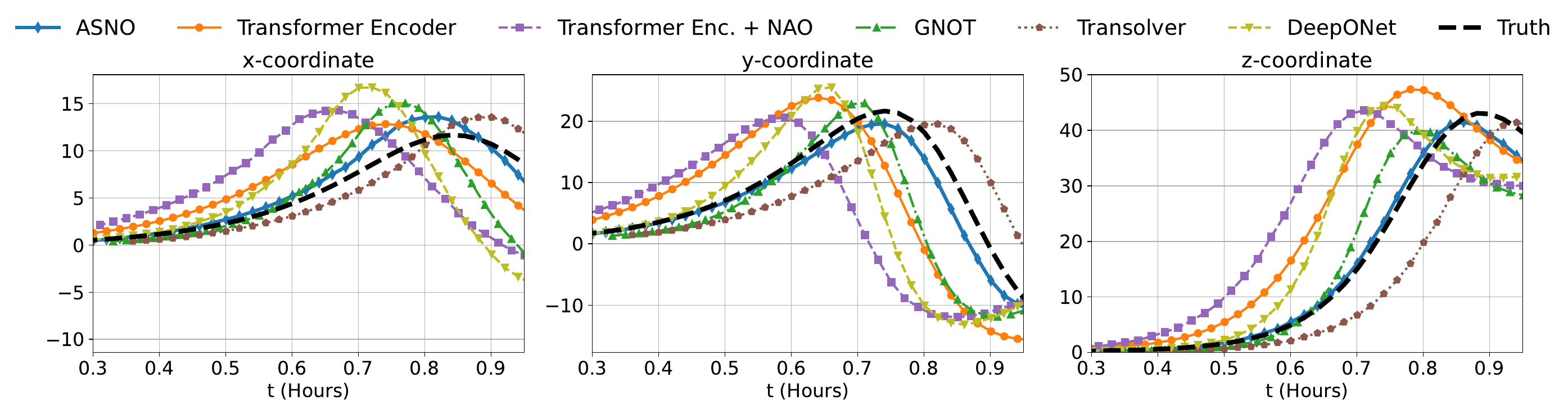}
    \vspace{-10pt}
    \caption{Long Time prediction comparison of Predicted vs. true time-series trajectories \((x, y, z)\) for models listed in Table~\ref{tab:chaotic_system_results} on the Lorenz system.}
    \label{fig:Lorenz_long}
\end{figure*}

\begin{table}[h!]
\caption{Comparison of ASNO and baselines models on the Lorenz system.}
\centering
\footnotesize
\begin{tabularx}{\linewidth}{>{\centering\arraybackslash}X
                             |>{\centering\arraybackslash}X
                             |>{\centering\arraybackslash}X
                             |>{\centering\arraybackslash}X}
\hline \hline
\textbf{Model Type} & 
\textbf{Trainable Params} & 
\textbf{GPU (MB)} & 
\textbf{Best Test Loss} \\
\hline
ASNO                      & 258,808                   & 76                     & \textbf{0.000794}       \\
\hline
Transolver                & 395,907                   & 95                     & 0.000835                \\
\hline
DeepONet                  & 265,512                   & 79                     & 0.001750                \\
\hline
Transformer Encoder       & \textbf{257,647}          & \textbf{71}            & 0.001821                \\
\hline
GNOT                      & 401,155                   & 106                     & 0.002189                \\
\hline
Linear Model + NAO        & 305,776                   & 87                     & 0.005298                \\
\hline \hline
\end{tabularx}
\label{tab:chaotic_system_results}
\end{table}

%ASNO uses a separable architecture that naturally aligns with how the Lorenz system is structured—splitting the dynamics into linear “homogeneous” terms and nonlinear interactions between variables:

By emulating an implicit–explicit BDF scheme, ASNO first employs a Transformer encoder to extrapolate the homogeneous (linear) dynamics from the past \(n\) states, thereby capturing dominant temporal modes and the system’s exponential divergence rate without interference from nonlinear feedback. The Nonlocal Attention Operator then applies a compact learned kernel to introduce the corrective coupling terms (e.g.\ the products \(xy\), \(xz\)), focusing exclusively on rapid cross-variable interactions. This two-stage decomposition lightens the representational load on each module: the Transformer Encoder solves a multistep linear recurrence, while the NAO specializes in a small nonlinear correction. As a result, ASNO achieves more stable long-term rollouts in the chaotic regime—small errors in one stage do not immediately amplify in the other—and empirical trajectories adhere much more closely to the true Lorenz attractor, especially in the sensitive \(y\) and \(z\) directions, than do those from monolithic models. For long-term rollouts, each new prediction is generated by feeding ASNO’s previous output back into the model as the next input, mirroring an autoregressive integration scheme. Figure~\ref{fig:Lorenz_long} shows ASNO’s predictions remaining near the ground truth over many steps, whereas the other baselines exhibit deviation as of earlier timesteps. By cleanly separating temporal forecasting from coupling correction, ASNO is able to achieve improved long-term stability and robustness—properties that are particularly desirable for chaotic forecasting tasks in climate science, physics simulations, and real-time control.

\subsection{Navier–Stokes Benchmark for PDE Testing}

To assess ASNO’s efficacy on complex PDEs, we evaluate it on the two-dimensional (2D) incompressible Navier–Stokes (NVS) equations in vorticity form. The governing equation is given by
\[
\frac{\partial \omega}{\partial t} + \mathbf{u} \cdot \nabla \omega = \nu \Delta \omega + f,\quad \text{with} \quad \Delta \psi = -\omega,\quad \mathbf{u} = \left(\frac{\partial \psi}{\partial y},\ -\frac{\partial \psi}{\partial x} \right),
\]
where \(\omega\) is the scalar vorticity in 2D, \(\mathbf{u} = (u, v)\) is the velocity field, \(\psi\) is the streamfunction, \(\nu\) is the viscosity, and \(f\) is the external forcing function. This benchmark tests ASNO’s ability to capture nonlinear interactions and multi-field dependencies, following the setup in \cite{li2020fourier}. The dataset consists of 100 simulated time-series profiles, each with a spatial resolution of \(30 \times 30\), governed by varying viscosity coefficients \(\nu \in [10^{-4}, 1]\), and subject to a forcing term–driven velocity field. Each profile contains 100 temporal snapshots with a time step size of \(\Delta t = 10^{-2}\). To construct training and testing data, a sliding window of length 5 and stride 1 is used, resulting in 96 samples per profile. Additionally, we apply 20 random permutations per trajectory to increase diversity and simulate realistic variations in flow dynamics. This yields \(80 \times 96 \times 20 = 153{,}600\) training samples and \(20 \times 96 \times 20 = 38{,}400\) testing samples. These settings create a comprehensive dataset for evaluating ASNO’s generalization and stability in modeling high-dimensional, nonlinear fluid systems.

\begin{figure*}[h!]
    \centering
    \includegraphics[width=\textwidth]{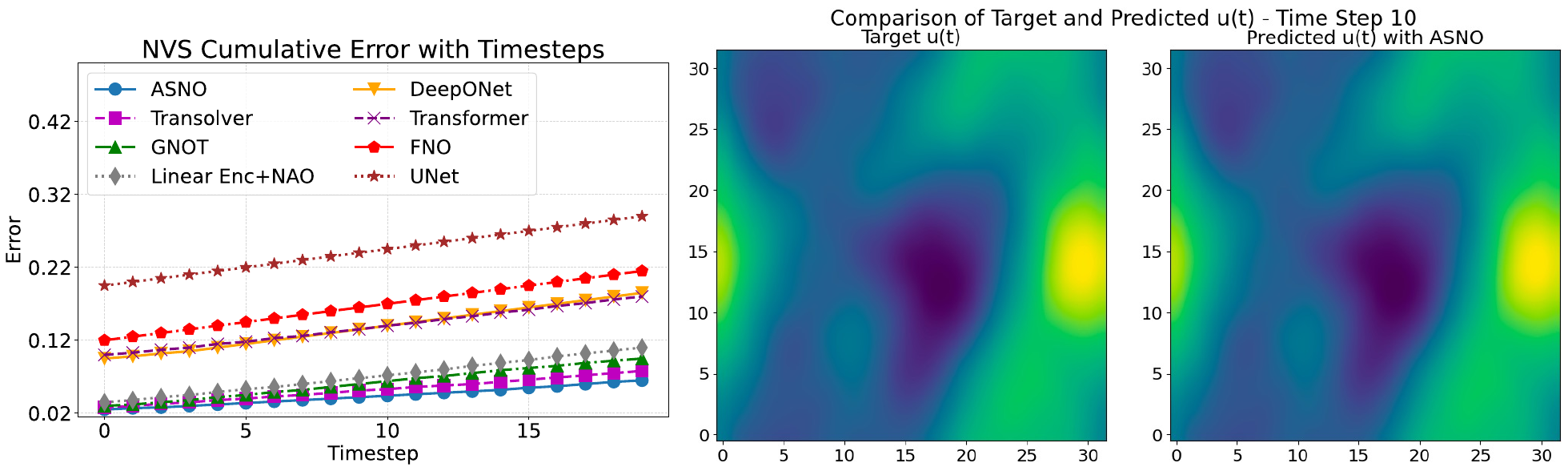}
    \vspace{-5mm}
    \caption{Left: Cumulative error comparison across models listed in Table~\ref{tab:nvs_results} on the Navier–Stokes benchmark. ASNO demonstrates minimal error growth over time, indicating long-term stability. Right: Visualization of predicted versus ground-truth velocity fields, showcasing ASNO’s ability to recover fine-scale flow structures.}
    \label{fig:NVS_cumulative_error}
\end{figure*}

\begin{table}[h!]
\centering
\footnotesize
\caption{Performance of models on Navier–Stokes Equations: Model Complexity, GPU Memory, and Best Test Loss.}
\label{tab:nvs_results}
\renewcommand{\arraystretch}{1.1} % row spacing
\setlength{\tabcolsep}{6pt}      % column spacing
\begin{tabularx}{\linewidth}{>{\centering\arraybackslash}X
                             |>{\centering\arraybackslash}X
                             |>{\centering\arraybackslash}X
                             |>{\centering\arraybackslash}X}
\hline\hline
\textbf{Model}   & \textbf{Trainable Params (M)} & \textbf{GPU (MB)} & \textbf{Best Test Loss} \\
\hline
ASNO        & 4.66  & 880   & \textbf{0.0213} \\
\hline
Transolver  & 4.14  & 911   & 0.0234         \\
\hline
GNOT        & 5.25  & 1024  & 0.0322         \\
\hline
DeepONet    & 5.10  & 3100  & 0.0921         \\
\hline
Trans. Enc. & 5.19  & 961   & 0.0967         \\
\hline
FNO        & 4.10  & 846   & 0.1186         \\
\hline
U-Net       & 5.02  & 991   & 0.1940         \\
\hline
Linear Encoder + NAO     & \textbf{4.05}  & \textbf{791}  & 0.0328 \\
\hline\hline
\end{tabularx}
\end{table}

Table~\ref{tab:nvs_results} compares ASNO against state-of-the-art models in test accuracy, efficiency, and parameter count. ASNO achieves the lower test loss while maintaining computational efficiency. Figure~\ref{fig:NVS_cumulative_error} illustrates cumulative error trends, demonstrating ASNO’s stability over long time horizons. These results confirm ASNO’s effectiveness in handling high-dimensional, nonlinear PDEs, generalizing across diverse physical systems with minimal numerical dissipation. Crucially, ASNO’s separable architecture—where the Transformer Encoder approximates the homogeneous (viscous and pressure) evolution through a high-order multistep forecast and the Nonlocal Attention Operator captures the advective and coupling interactions spatially—reduces the representational burden on each component. By isolating the stiff linear dynamics from the convective nonlinearities, ASNO attains both greater numerical stability and sharper resolution of fine-scale flow structures than monolithic models. Building on ASNO’s success in modeling high-dimensional, nonlinear PDEs like the Navier–Stokes equations, we extend its application to real-world engineering challenges.

\subsection{Melt Pool Temperature Field Prediction in Additive Manufacturing}

In this section we present a case study on Directed Energy Deposition (DED) AM processes where the goal is to predict the evolution of the melt pool temperature field \cite{gunasegaram2024machine}. Accurate melt pool prediction is crucial in real-world manufacturing scenarios as small deviations in melt pool temperature can significantly affect part quality. First, to collect data to train ASNO we employ GAMMA, an in-house developed GPU-accelerated Finite Element Analysis (FEA) code, for part-scale simulations \cite{liao2023efficient}. The GAMMA simulation follows the transient heat conduction equation, incorporating the essential partial differential equations (PDEs) that underpin the DED process:
\begin{equation} 
\rho C_p(T)\frac{\partial T}{\partial t} + \nabla\cdot q = 0,  
\end{equation}
Where $T$ is the temperature (K), $\rho$ and $C_p$ are the density ($g/mm^3$) and the effective specific heat capacity ($J/g/K$) of the material, respectively. The heat flux $q$ is given by Fourier's law:
\begin{equation}
    q = -k \nabla T, 
\end{equation}
where $k$ is the thermal conductivity of the material. The boundary conditions in the DED process can be formulated as:
\begin{equation}
\begin{aligned}
    q\cdot n = \frac{-2\eta P}{\pi r_{beam}^2} \exp\left(\frac{-2d^2}{r_{beam}^2}\right) + \sigma \epsilon (T^4 - T_0^4) + h(T - T_0), 
\end{aligned}
\end{equation}
\noindent where $\eta$ is the absorption coefficient (\%), $P$ is the laser power (W), $r_{beam}$ is the beam radius (mm), and $d$ is the distance (mm) from the material point to the center of the laser.  $h$ is the convection heat transfer coefficient ($W/m^2/K$), $\sigma$ is the Stefan-Boltzmann constant ($5.67 \times 10^{-8} W/m^2/K^4$), $\epsilon$ is the material's emissivity, and $T_0$ is the ambient temperature. In the simulation, elements are activated and incorporated into the mesh when the distance between the element’s center and the beam center is less than the beam’s size.

\begin{figure}[h!]
    \centering
    \includegraphics[width=1\linewidth, trim={10 0 10 10}, clip]{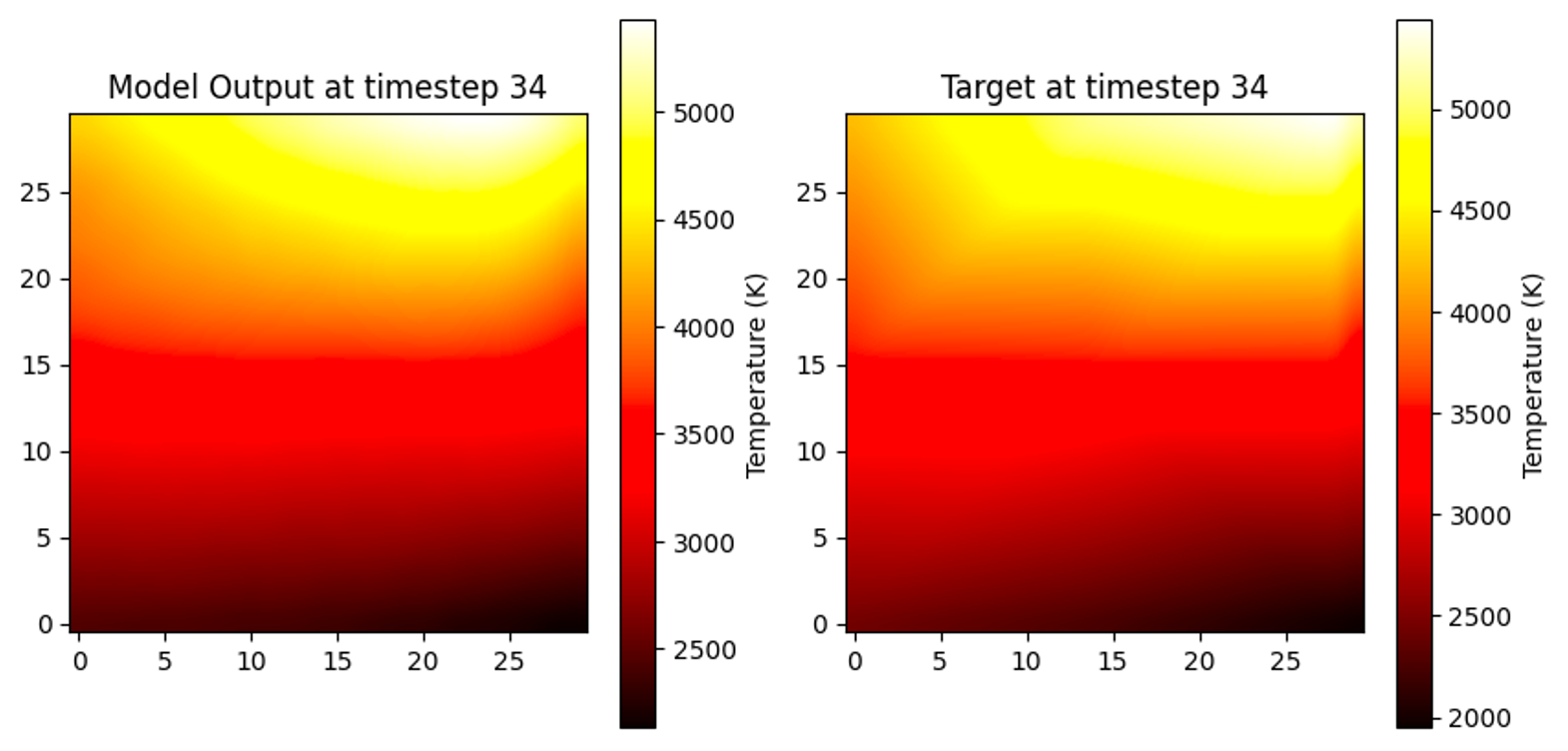}  % Adjust the trim values as needed
    \vspace{-0.5cm}
    \caption{Prediction results on the DED AM case study with the ASNO prediction on the left, and target results on the right.}
    \label{fig:AM_Results}
\end{figure}

To generate training data, we first generated 100 different laser power profiles with various combinations of laser power $P_{\text{laser}}(m)$ and scanning rate $V_{\text{scan}}(m)$ as process parameters, and then simulate the printing process of a thin wall using the GAMMA simulation. The temperature field $T_{\text{pool}}(m)$ corresponding to the laser location $L_{\text{laser}}(m)$ at each time step $m$ was saved as the training output. Each temperature field is recorded on a $21 \times 21$ spatial grid, and each profile consists of 730 time steps. To construct the training data, a sliding window of length 5 and stride 1 was used, resulting in $730 - 5 + 1 = 726$ samples per profile. A random permutation of 20 per trajectory was applied to enhance data diversity while preserving temporal locality, yielding a total of $80 \times 726 \times 20 = 1{,}161{,}600$ training samples and $20 \times 726 \times 20 = 290{,}400$ testing samples (based on an 80:20 split). With training data collected, the ASNO for predicting the melt pool temperature field can be trained by formulating the prediction based on the processing parameters and the history from the past five steps:

\begin{equation}
\begin{aligned}
    \hat{T}_{pool}(m+1) &= \text{ASNO}\left( P_{laser}(m+1), V_{scan}(m+1), \right. \\
    &\quad \left. L_{laser}(m+1), T_{pool}(m-4:m) \right).
\end{aligned}
\end{equation}
The predicted melt pool temperature field from a selected frame is shown in Figure \ref{fig:AM_Results}, compared with the ground truth result from the GAMMA simulation. As a result, the ASNO achieves a Mean Absolute Percentage Error (MAPE) of 2.50\%. Additional prediction results of different process input scenarios for additive manufacturing are provided in Appendix~A.

% \begin{figure}[h!] % 'b' specifies the bottom of the page
%     %\vspace{-0.5cm}  % Adjust the vertical spacing above the figure if needed
%     \includegraphics[width=1\linewidth]{AM_result.png}  % Replace with the actual path of your image file
%     \vspace{-0.5cm}  % Reduce the vertical spacing below the figure for a tighter fit
%     \caption{Prediction results on the DED case study with the ASNO results on the left, and target results on the right.}
%     \label{fig:AM_Results}
%  % Adjust this value as needed to reduce space above the table

% \end{figure}

\section{Conclusion}

In this paper, we introduce ASNO, a spatio-temporal framework designed to enhance generalizability and interpretability in modeling time-dependent differential equations. ASNO leverages a separable kernel approach inspired by the implicit-explicit BDF, enabling to disentangle temporal and spatial interactions to provide insights on system dynamics. The Transformer Encoder estimates homogeneous system solutions from historical states, analogous to the explicit step in BDF, while a neural operator models spatial interactions and external load impacts. To ensure robust generalization across diverse physical systems, ASNO integrates the NAO, which facilitates adaptation to varying PDE parameters such as initial conditions, loadings, and environments. Evaluations on benchmarks including the Lorenz system, Darcy flow equation, Navier Stokes, and the DED application demonstrate that ASNO outpertforms over the existing models in terms of accuracy, long-term stability, and zero-shot generalizability to unseen physical parameters. In addition, ASNO offers interpretability by isolating the influence of temporal and spatial dynamics, revealing how historical states and spatial loadings contribute to governing behaviors, and aligning predictions with underlying physical laws. This dual emphasis on generalizability and interpretability makes ASNO suitable for real-time decision-making in high-stakes environments, while its attention mechanisms uncover meaningful patterns for physics-informed discovery and decision-making. Future research can extend ASNO’s applications to foundational modeling for transfer learning across different physical systems, further advancing its role in uncovering and predicting complex physical phenomena in scientific and industrial domains.

\section*{Acknowledgements}
We appreciate the grant support from the NSF HAMMER-ERC (Engineering Research Center for Hybrid Autonomous Manufacturing, Moving from Evolution to Revolution) under Award Number EEC-2133630, and the NSF MADE-PUBLIC Future Manufacturing Research Grant Program under Award Number CMMI-2037026. Wei Chen and Doksoo Lee acknowledge support from the NSF Boosting Research Ideas for Transformative and Equitable Advances in Engineering (BRITE) Fellow Program (CMMI 2227641). Yi-Ping Chen appreciates the Taiwan-Northwestern Doctoral Scholarship funded by the Ministry of Education in Taiwan, and the fellowship support from the Predictive Science and Engineering Design (PSED) cluster at Northwestern University. Yue Yu was funded by the National Institute of Health under Award 1R01GM157589-01 and the AFOSR under Grant FA9550-22-1-0197. Portions of this research were conducted on Lehigh University’s Research Computing infrastructure, partially supported by NSF Award 2019035.

\bibliography{example_paper}
\bibliographystyle{apalike}

\newpage
\appendix
\onecolumn
\icmltitle{Appendix}

% \appendix

% \section*{\centering \Large \bfseries Appendix}

\begingroup
% \small % Reduces font size for the appendix
\setlength{\textfloatsep}{1pt} % Reduces space between text and figures

\section{Melt Pool Temperature Field Prediction in Additive Manufacturing}

\begin{figure}[H]
    \centering
    \includegraphics[width=1\linewidth, height=0.75\textheight, keepaspectratio]{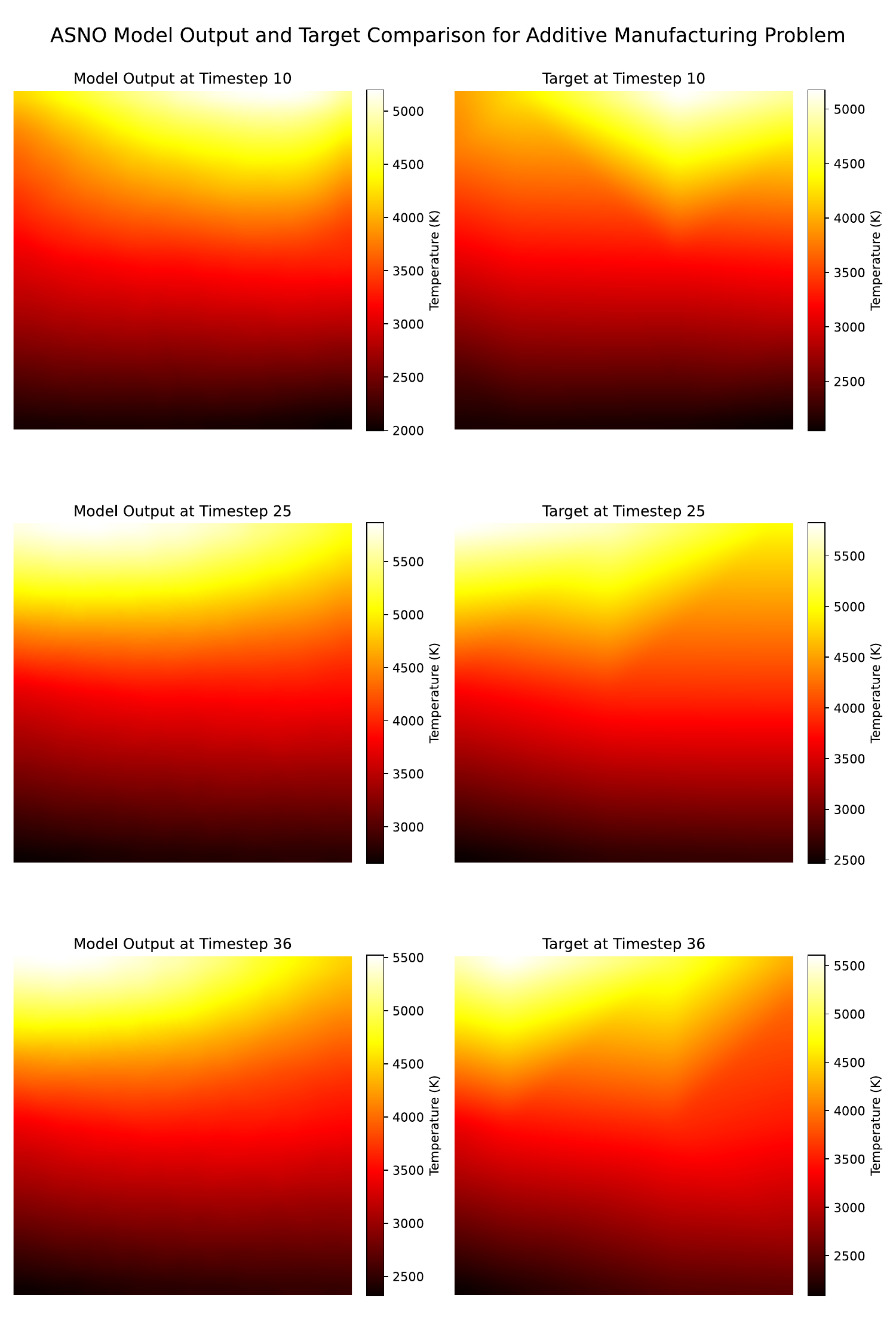}
    \caption{Comparison of ASNO model output and target temperature distributions for the melt pool in Directed Energy Deposition at Timesteps 10, 25, and 36. Each row represents a different timestep, with the model output shown on the left and the target on the right. The color scale indicates temperature in Kelvin, highlighting ASNO's accuracy in capturing the thermal profile of the melt pool.}
    \label{fig:meltpool_results}
\end{figure}
% \begin{figure}[H]
%     \centering
%     \includesvg[width=1\linewidth, height=0.75\textheight, keepaspectratio]{model_output_comparison_multiple_timestep1} % Replace with the actual file name
%     \caption{Comparison of ASNO model output and target temperature distributions for the melt pool in Directed Energy Deposition at Timesteps 10, 25, and 36. Each row represents a different timestep, with the model output shown on the left and the target on the right. The color scale indicates temperature in Kelvin, highlighting ASNO's accuracy in capturing the thermal profile of the melt pool.}
%     \label{fig:meltpool_results}
% \end{figure}

In this section, we provide a comprehensive analysis of the melt pool characteristics predicted by ASNO and other comparative models. The melt pool in Directed Energy Deposition (DED) is a key factor in determining material properties. Accurate modeling of the melt pool helps assess temperature distribution, phase transformation, and thermal cycles, which directly affect the microstructure and mechanical performance of the final product. Additionally, understanding melt pool dynamics is essential for process control, defect mitigation, and ensuring part consistency in DED applications. By comparing ASNO’s predictive performance with other models, we aim to demonstrate how well it captures these critical characteristics in a manufacturing setting.

The simulations were conducted over a \(21 \times 21\) spatial grid, with temperature distributions captured at multiple timesteps throughout the deposition process. In each timestep, the model outputs a 2D temperature field that reflects the temperature variations across the substrate due to the moving heat source. 

Figure \ref{fig:meltpool_results} illustrates these results, showing a side-by-side comparison between ASNO's model predictions and the target values across three distinct timesteps (10, 25, and 36). The color gradient, representing temperature from 2000 K to 5500 K, demonstrates the intensity of heat within the melt pool and the cooling pattern radiating outward from the laser’s path. In each row, the predicted temperature distribution is shown on the left, directly alongside the corresponding target temperature distribution on the right, offering a clear visual comparison of ASNO's accuracy at various time steps. The ASNO results show good agreement with the target profiles. 

These visualizations underscore ASNO’s ability to model high-temperature zones and predict the progression of thermal profiles over time accurately. Notably, the model predictions retain stability over longer timesteps, capturing the complexity of the transient heat flow in DED processes. The agreement between the predicted and target distributions suggests that ASNO is well-suited for high-fidelity simulations in additive manufacturing applications, where precise thermal predictions are crucial for quality control and optimization.

\section*{Nomenclature}
\begin{table}[h!]
\centering
\footnotesize
\setlength{\tabcolsep}{6pt}
\begin{tabularx}{\linewidth}{@{}lX@{}}
\toprule
Symbol & Meaning \\
\midrule
$X,\;x$                                   & State vector (general) \\
$X_m$                                     & State at timestep $m$ \\
$\tilde{X}_{m+1}$                         & Extrapolated (homogeneous) state \\
$F$                                       & External forcing / loading field \\
$S$                                       & Hidden system (environment) state \\
$t$                                       & Continuous time \\
$m$                                       & Time‐index \\
$n$                                       & History length \\
$\Delta t$                                & Time step size \\
$\alpha_k,\;\beta$                        & BDF coefficients (see \eqref{eq:BDF}) \\
$\mathcal{D}^{(\eta)}$                    & Dataset for system $\eta$ \\
$\mathcal{F}$                             & ASNO operator mapping \\
$\hat y_{t+1}$                            & Predicted output at time $t+1$ \\
$y_t,\;\hat y_t$                          & True and predicted values at time $t$ \\
$E_T$                                     & Cumulative error over $T$ steps \\
$e_t$                                     & Instantaneous error at timestep $t$ \\
$Q,\,K,\,V$                               & Query, Key, and Value matrices used in attention mechanisms; used generically, with context-specific definitions below \\
$W_E$                                     & Embedding matrix \\
$W_{Tq},\,W_{Tk},\,W_{Tv}$                & Time‐series Transformer Encoder attention weight matrices for Query ($Q$), Key ($K$), and Value ($V$) used in the explicit extrapolation step \\
$P_k$                                     & Positional encoding at position $k$ \\
$d_{\mathrm{embed}},\,d_t,\,d$            & Embedding dimension, key/query dimension, and feature dimension \\
$H_t$                                     & Penultimate‐step latent features used in computing the NAO kernel weights for $h$ and $f$ \\
$J_t$                                     & Intermediate NAO state at attention step $t$ composed of latent features and forcing fields $(H_t, F_t)$, iteratively updated across $T$ layers \\
$W_{P,h},\,W_{P,f},\,W_{Q_t},\,W_{K_t}$   & Weight matrices in the Nonlocal Attention Operator (NAO) used for computing kernel projections; $W_{Q_t}$ and $W_{K_t}$ generate attention Query and Key vectors used in the implicit interaction modeling \\
$K[\cdot]$                                & Learned nonlocal kernel operator acting over latent space in NAO \\
$\mathcal{H},\,\mathcal{F}$               & Input and output Banach function spaces \\
$T$                                       & Number of attention steps in NAO \\
\bottomrule
\end{tabularx}
\end{table}

\end{document}